%% file: main.tex
\documentclass[conference]{IEEEtran}
\IEEEoverridecommandlockouts

\usepackage{amsmath,amssymb,amsfonts}
\usepackage{multirow}
\usepackage{xcolor}
\usepackage{tikz}
\usepackage{booktabs}
\usepackage{algorithm}
\usepackage{graphicx}
\usepackage{microtype}
\usepackage{textcomp}
\usepackage{xcolor}
\usepackage[noend]{algpseudocode}
\usepackage[
   backend=biber,
   style=ieee,
   citestyle=numeric-comp,
   maxbibnames=5,
   maxcitenames=5,
   doi=false,isbn=false,url=false,eprint=false
]{biblatex}
\input{sourcemap_definitions.tex}
\defbibheading{bibliography}[\refname]{}

\begin{document}

\title{All In Good Time: Causality-Aware Framework for LLM-Based Simultaneous Speech-to-Speech Translation
}

\author{
\IEEEauthorblockN{Amir Hussein\IEEEauthorrefmark{1}\IEEEauthorrefmark{2},
Enas Albasiri\IEEEauthorrefmark{2},
Travis M. Bartley\IEEEauthorrefmark{2},
Nourchene Ferchichi\IEEEauthorrefmark{2},
Ke Hu\IEEEauthorrefmark{2},
Harishchandra Dubey\IEEEauthorrefmark{2},\\
Myungjong Kim\IEEEauthorrefmark{2},
Zhehuai Chen\IEEEauthorrefmark{2},
Oluwatobi Olabiyi\IEEEauthorrefmark{2}, Sanjeev Khudanpur\IEEEauthorrefmark{1}}
\IEEEauthorblockA{
\IEEEauthorrefmark{1}Johns Hopkins University,
\IEEEauthorrefmark{2}NVIDIA
}
\IEEEauthorblockA{\texttt{amhussein@nvidia.com}}
}

\maketitle

\begin{tikzpicture}[remember picture,overlay]
\node[anchor=south, yshift=6pt] at (current page.south) {%
  \parbox{0.9\paperwidth}{\centering\scriptsize
    \textcopyright{} 2026 IEEE. Personal use of this material is permitted.
    Permission from IEEE must be obtained for all other uses, in any current
    or future media, including reprinting/republishing this material for
    advertising or promotional purposes, creating new collective works,
    for resale or redistribution to servers or lists, or reuse of any
    copyrighted component of this work in other works.
  }%
};
\end{tikzpicture}

\begin{abstract}

Large Language Models (LLMs) have shown strong performance in low-resource offline translation; however, extending them to simultaneous speech-to-speech translation (Simul-S2ST) remains challenging due to the scarcity of causally aligned training data with high cross-lingual speaker fidelity. In addition, existing approaches rely on fixed translation policy or confidence heuristics, leading to suboptimal quality and higher latency. We propose a causality-aware Simul-S2ST framework with a novel data pipeline that generates high-fidelity, causally aligned segments with improved voice transfer. The framework introduces (i) a factorized S2ST architecture (FAST), (ii) a causality-aware adaptive policy (CAP), and (iii) causality-aware latency metric. Experiments on CVSS Spanish, German, and French show that FAST-CAP consistently improves the quality--latency trade-off, achieving up to +1.2 BLEU and a 26\% relative latency reduction over a fixed policy. Despite using substantially less training data than existing systems, FAST-CAP achieves state-of-the-art results in speech translation quality and speaker fidelity while yielding up to a 38.8\% relative reduction in latency.
\end{abstract}


\section{Introduction}
Simultaneous speech-to-speech translation (Simul-S2ST) aims to translate speech from one language into another in real time, allowing natural cross-lingual interactions. Beyond low latency, human-like dialogue requires cross-lingual voice transfer that preserves speaker identity while appropriately transferring paralinguistic cues across languages~\cite{xie2026prosody}. It must also handle conversational phenomena such as overlapping speech and interruptions, making Simul-S2ST particularly challenging~\cite{ahmad-etal-2024-findings}. Another major challenge is the scarcity of paired S2ST training data~\cite{agrawal-etal-2023-findings, hussein-etal-2023-jhu}, especially compared to the abundance of transcribed speech and translated text. This disparity motivates a key question: \textbf{How can Simul-S2ST systems leverage large pretrained models trained on abundant speech and text data?} 

Recently, there has been growing interest in leveraging pretrained large language models (LLMs) as translation backbones, motivated by their strong contextual reasoning, long-context modeling capabilities, and improved translation quality in low-resource and zero-shot settings~\cite{huang2023speech_translation_llm,survey_llm_mt_2025,lyu-etal-2024-paradigm,koshkin-etal-2024-transllama}. These advances have led to increased interest in LLM-based simultaneous translation to exploit sequence modeling capacity of LLMs for low-latency translation. However, existing LLM-based simultaneous translation approaches largely rely on fixed or heuristic read/write policies, such as wait-$k$, which may generate output before sufficient source context is available and degrade translation quality~\cite{agostinelli2024simul,koshkin2024llms,futami25_interspeech,ouyang2025infinisst}. Moreover, most speech translation systems focus on lexical content and overlook rich paralinguistic cues such as prosody, emotion, attitude, and speaker intent~\cite{jia2022translatotron,barrault2023seamless}.

To jointly optimize speech perception and generation, recent work has explored direct Simul-S2ST models that translate source speech into target speech while preserving expressive speech cues~\cite{barrault2023seamless}. Hibiki~\cite{labiausse2025highfidelity} extends this direction with an LLM-based multi-stream architecture and a dynamic translation policy driven by offline MT perplexity. However, offline MT perplexity tends to favor longer source contexts, leading to high latency. Moreover, using a single neural codec representation for both speech perception and generation introduces competing objectives and can result in suboptimal performance~\cite{hussein2025hasrd}.

To address the aforementioned limitations, we introduce a causality-aware framework\footnote{Code: \url{ https://github.com/AmirHussein96/FAST-CAP/tree/main.}} for LLM-based Simul-S2ST. LLM-based Simul-S2ST requires aligned speech-to-speech data to learn when to wait and when to speak, but such data is unavailable. To bridge this gap, we develop a unified data pipeline that constructs causally aligned training examples and synthesizes natural target speech with high-fidelity cross-lingual voice transfer. Inspired by cognitive studies on chunking strategies used by professional interpreters~\cite{huang2023chunking, song2023aptitude}, we introduce a causality-aware adaptive policy in which read/write decisions are guided by the availability of sufficient source information to generate the corresponding translation, resulting in optimal waiting strategy. To mitigate the limitations of a shared codec representation, we propose a factorized architecture (FAST) that decouples lexical and acoustic modeling: lexical information is extracted by an ASR encoder, while acoustic information is modeled using neural audio codec. FAST adopts a multi-stream full-duplex design~\cite{hu2025efficient} that enables effective handling of multimodal inputs, overlapping speech, and conversational interruptions. Finally, we propose CAAL, an alignment-aware latency metric that measures only avoidable delay beyond the ideal causal policy. Our key contributions are:  
\begin{itemize}
    \item A data generation pipeline with causality-aware adaptive policy (CAP).
    \item A factorized architecture (FAST) that decouples lexical and acoustic representations while preserving speaker identity and vocal characteristics.
    \item A causality-aware latency metric (CAAL) that distinguishes necessary linguistic delays from avoidable system delays.
\end{itemize}

\section{Proposed Approach}

Our proposed factorized simultaneous speech-to-speech translation (FAST) model uses an LLM as the backbone for translation and adopts a multi-stream full-duplex design~\cite{hu2025efficient}. The multi-stream design enables real-time processing different modalities and overlapping speech as separate but temporally aligned streams.
\begin{table}[tb!]
    \centering
    \vspace{-0.4cm}
    \caption{Comparison of speaker similarity and audio quality on the CVSS-T Dataset, with audio quality measured by UTMOS-V2.}
    \include{tables/data}
    \vspace{-0.4cm}
    \label{tab:data}
\end{table}
\subsection{Data Generation Pipeline}\label{sec:data_pipeline}
\begin{figure}[tb!]
    \centering
   \vspace{-0.4cm}\includegraphics[width=1\columnwidth]{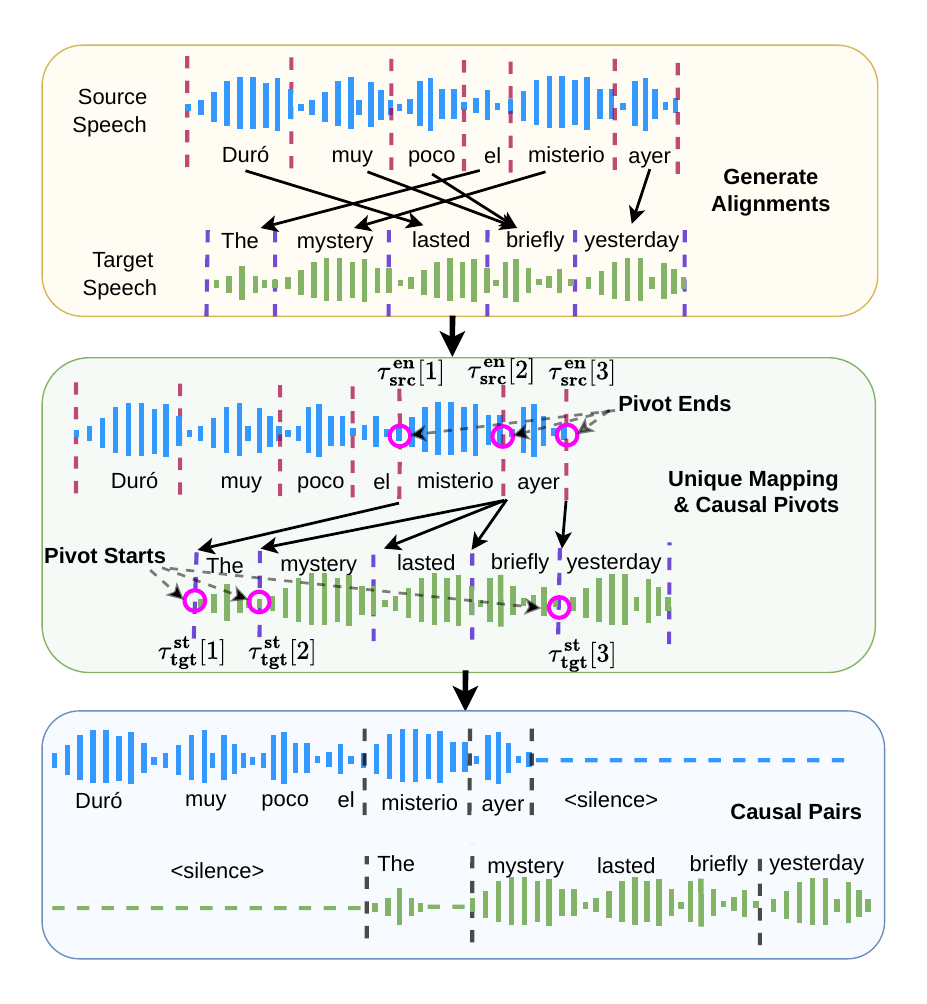}
    
    \caption{Overview of the data generation pipeline. The process includes (1) alignments generation, (2) unique mapping and causal pivot extraction, and (3) causality-aware adaptive chunks construction.}
    \label{fig:data_pipeline}
    \vspace{-0.4cm}
\end{figure}

To enable simultaneous processing, the generated translations must be causal emitting outputs as soon as sufficient source information becomes available with minimal latency. 
Figure~\ref{fig:data_pipeline} illustrates the proposed data generation pipeline with causality-aware adaptive chunking.
\\
\textbf{Improving cross-lingual speaker fidelity:} Speech-to-speech translation models are commonly trained on paired data with target speech synthesized using cross-lingual TTS, as in CVSS-T~\cite{jia2022cvss}. However, CVSS-T exhibits low source--target speaker similarity, with an average ECAPA-TDNN cosine similarity of 0.24 as shown in Table~\ref{tab:data}. We therefore resynthesize CVSS-T using the zero-shot TTS model A2Flow\footnote{\url{https://catalog.ngc.nvidia.com/orgs/nvidia/teams/nvigisdk/models/riva-tts-a2flow}}, which increases the average speaker similarity to 0.61.\\
\textbf{Alignment generation:} Training an LLM-based Simul-S2ST system requires causally aligned source--target pairs so the model can learn when to wait and when to generate translation. We first align the source speech $\mathbf{s}^{src}_{1:T}$ with its transcript $\mathbf{f}=(f_1,\dots,f_J)$ using Montreal Forced Aligner (MFA)~\cite{mcauliffe2017montreal}. We then use Awesome-align~\cite{dou2021word} with multilingual BERT to obtain word alignments between $\mathbf{f}$ and the target translation $\mathbf{e}=(e_1,\dots,e_I)$. Following the IBM alignment formulation, we define the raw alignment set as $\mathcal{A} = \{(j,i) \mid j \in \{1,\dots,J\},\ i \in \{1,\dots,I\}\}$, where $(j,i)$ indicates that source word $f_j$ is aligned to target word $e_i$; see Figure~\ref{fig:data_pipeline}.\\
\textbf{Causality-Aware Adaptive Policy:} 
We remove crossing dependencies by converting the raw text-to-text alignment into a monotonic alignment $\Pi$. 
This is non-trivial due to word-order differences, many-to-one alignments and omitted words in translation. To address these challenges, Algorithm~\ref{alg:monotonic_alignment} introduces \textsc{MonotonicAlign}$(\cdot)$, which builds a monotonic target-to-source alignment by sorting alignments by target index, retaining the highest source index for each target word, and enforcing non-decreasing source indices with a running maximum. Missing target alignments are then filled with a backward pass. The resulting pivot alignments $\mathcal{P}$ serve as anchors for causal segmentation, as shown in Figure~\ref{fig:data_pipeline}. We optionally apply \textsc{MergeChunks}$(\cdot)$ to merge adjacent short chunks into longer causal segments. The \textsc{MonotonicAlign}$(\cdot)$ function returns a set of pivot alignments with unique indices, denoted by $\mathcal{P}$, which serve as anchors for causal segmentation. The behavior of $\textsc{MonotonicAlign}(\cdot)$ is illustrated in the \emph{Unique Mapping \& Causal Pivots} step of Figure~\ref{fig:data_pipeline}. We then optionally apply $\textsc{Combine}(\cdot)$ to merge adjacent causal chunks into longer causal segments, analogous to phrase-based translation from statistical machine translation. Finally, the causal emission policy is implemented through $\textsc{CausalPairedChunks}(\cdot)$, which constructs causal source-target chunks around these pivots. On the source side, each chunk spans from the end of the previous pivot to the end of the current pivot (left aggregation). On the target side, each chunk spans from the current pivot’s target start to the next pivot’s target start (right aggregation). This construction ensures each target token is emitted only after the necessary source evidence has been observed, forming a Causality-Aware Adaptive Policy (CAP) for Simul-S2ST training. Formally, for each target token $e_i$, let $R(i) := \{ j \mid (j,i) \in \Pi \}$ denote the source indices required to generate $e_i$ and and let $t^{en}_{src}[j]$ denote the end time of source word $f_j$. We define the source-side pivot end time $\tau^{en}_{src}[i]$ recursively as:
\begin{equation}
\tau^{en}_{src}[i] = \max \left( \tau^{en}_{src}[i-1],\ \max_{j \in R(i)} t^{en}_{src}[j] \right).
\end{equation}
Generation is therefore restricted to the observed speech prefix, such that $P(e_i \mid \mathbf{s}_{1:T}) = P(e_i \mid \mathbf{s}_{1:\tau^{en}_{src}[i] })$. This ensures causal generation with minimal latency. When a target speech chunk (e.g., ``The'') is shorter than the next aligned source segment (e.g., ``misterio''), we append silence to preserve causal timing. To reduce boundary artifacts, silence gaps are smoothed using mirrored edges from neighboring chunks with a Hamming window. Alignment processing and causal chunk construction are implemented with Lhotse~\cite{zelasko2021lhotse}.
\begin{algorithm}[tb!]
\caption{Causality-Aware Adaptive Policy}
\label{alg:monotonic_alignment}
\scriptsize
\begin{algorithmic}[1]

\Require Source words $\mathbf{f}=(f_1,\dots,f_J)$, $\mathbf{t}_{src}=\{(t_{j}^{\mathrm{st}}, t_{j}^{\mathrm{en}})\}_{j=1}^{J}$ 
\Require  Target words $\mathbf{e}=(e_1,\dots,e_I)$,  $\mathbf{t}_{tgt}=\{(t_{i}^{\mathrm{st}}, t_{i}^{\mathrm{en}})\}_{i=1}^{I}$
\Require  Text alignments $\mathcal{A} \subseteq \{1..J\}\times\{1..I\}$ 
\Require  Minimum source chunk duration $d_{\min}$
\Function{MonotonicAlign}{$\mathcal{A}, \mathbf{f}, \mathbf{e}$}
    \State $\Pi \gets \textsc{Sort}(\mathcal{A}, \text{by target index } i, \text{then source index j})$
    \State Retain highest source index for each target index
    \For{$k = 2$ to $|\Pi|$}
    \State $\pi_k[src] \gets \max(\pi_k[src], \pi_{k-1}[src])$ \Comment{enforce monotonicity}
    \EndFor
    \For{$i = I-1$ to $1$} \Comment{loop over target indices}
    \If{$\nexists\ \pi \in \Pi:\pi[tgt]=i$} 
    \State $\Pi \gets \Pi \cup \{(\pi_{i+1}[src],i)\}$ \Comment{add missing targets}
    \EndIf
    \EndFor
    \State $\mathcal{P} \gets \mathrm{GetPivots(\Pi)}$ 
    \Comment{extract pivots for causal chunks}
    \State \Return $\mathcal{P}$
\EndFunction


\Function{CausalPairedChunks}{$\mathcal{P}^*$, $\mathbf{t}_{src}, \mathbf{t}_{tgt}, \mathbf{f}, \mathbf{e}$}
    \State $C^{\text{src}}, C^{\text{tgt}} =\{\emptyset\}_{k=1}^{|\mathcal{P}^*|}$;  \Comment{initialize source and target chunks dict}
    \State $\tau^{en}_{src} \gets $\textsc{PivotSrcEn($\mathcal{P}^*[src]$, $\mathbf{t}_{src}$)} \Comment {get pivot src end time}
    \State $\tau^{st}_{tgt} \gets $\textsc{PivotTgtSt($\mathcal{P}^*[tgt]$, $\mathbf{t}_{tgt}$)} \Comment {get pivot tgt start time}
    \For{$r = 1$ to $|\mathcal{P}^*|$} \Comment{iterate over pivot indices}
    \State $C_r^{src} \gets$ [$\mathbf{f}_{r-1:r}$, $(\tau^{en}_{src}[r-1],\tau^{en}_{src}[r])$] \Comment{src left aggregation}
     \State $C_r^{tgt} \gets$ [$\mathbf{e}_{r:r+1}$, $(\tau^{st}_{tgt}[r],\tau^{st}_{tgt}[r+1])$] \Comment{tgt right aggregation}
    \EndFor
    \State \Return $C^{src}$,$C^{tgt}$
\EndFunction

\State $\mathcal{P} \gets$ \textsc{MonotonicAlign}$(\mathcal{A}, \mathbf{f}, \mathbf{e})$
\State $\mathcal{P}^*\gets$ \textsc{MergeChunks}$(\mathcal{P}, d_{min})$
\State $C^{src}, C^{tgt} \gets$ \textsc{CausalPairedChunks}$(\mathcal{P}^*,\mathbf{t}_{src}, \mathbf{t}_{tgt},\mathbf{f}, \mathbf{e})$
\State \Return $C^{src},C^{tgt}$
\end{algorithmic}
\end{algorithm}

\subsection{Model Architecture}
\begin{figure}[tb!]
    \centering
    \vspace{-0.4cm}\includegraphics[width=1\columnwidth]{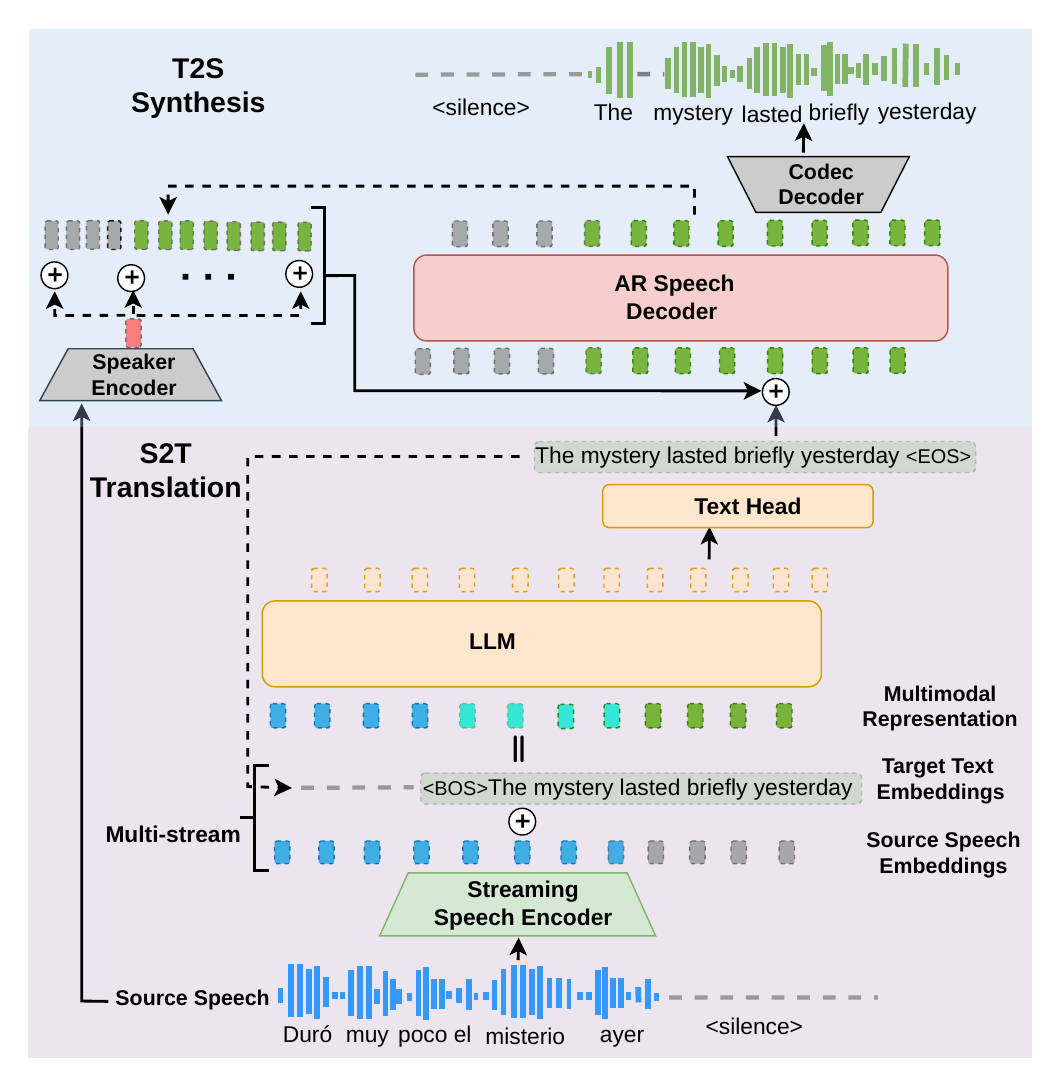}
    
    \caption{Overview of the proposed FAST architecture with multistream joint sequence modeling.}
    \vspace{-0.4cm}
    \label{fig:architecture}
\end{figure}

The FAST architecture adopts a multi-stream design~\cite{hu2025efficient}, as illustrated in Figure~\ref{fig:architecture}. Unlike prior work~\cite{labiausse2025highfidelity}, which uses a single neural codec representation for both perception and generation, FAST decouples lexical and acoustic representations inspired by~\cite{hussein25_interspeech}. Lexical information is provided to the LLM through an ASR encoder, while acoustic information is modeled by an autoregressive TTS module using codec tokens. This design improves translation accuracy while preserving high-quality speech synthesis. To enable cross-lingual voice transfer, we extract source-speaker embeddings using TitaNet\footnote{\url{https://huggingface.co/nvidia/speakerverification_en_titanet_large}} and add them to the codec-token representations. Speech is tokenized with streaming NanoCodec~\cite{casanova25_interspeech}, which uses Finite Scalar Quantization (FSQ). Unlike RVQ-based codecs~\cite{labiausse2025highfidelity}, FSQ uses independent codebooks, enabling parallel codebook prediction at each timestep. Given a causally aligned segment from Section~\ref{sec:data_pipeline}, let $\mathbf{Q}=(\mathbf{q}_1,\dots,\mathbf{q}_K)$ denote the target speech codec sequence, where each frame contains $N_q$ discrete code indices, $\mathbf{q}_k \in \{1,\dots,|\mathcal{V}^{(q)}|\}^{N_q}$. The target text sequence $\mathbf{e}=(e_1,\dots,e_i,\langle\text{pad}\rangle)$ is padded to match the codec length, such that $|\mathbf{Q}|=|\mathbf{e}|$. FAST processes source speech prefix $\mathbf{s}_i:=\mathbf{s}_{1:\tau^{en}_{src}[i]}$ incrementally and generates both text tokens $\mathbf{e}$ and codec tokens $\mathbf{Q}$ in a streaming fashion. We factorize the joint conditional distribution $P(\mathbf{Q}, \mathbf{e} \mid \mathbf{s})$ as
\begin{equation}
\begin{aligned}
P(\mathbf{Q}, \mathbf{e} \mid \mathbf{s})
&= P(\mathbf{e} \mid \mathbf{s})
   \; P(\mathbf{Q} \mid \mathbf{e}, \mathbf{s}) 
\end{aligned}
\end{equation}
where $P(\mathbf{e} \mid \mathbf{s})$ is the speech to text translation module (Simul-S2T), and $P(\mathbf{Q} \mid \mathbf{e}, \mathbf{s})$ is the text-to-speech synthesis module (Simul-T2S). FAST parameterizes $P(\mathbf{e} \mid \mathbf{s})$ with an ASR-pretrained streaming speech encoder, followed by a pretrained LLM. The speech encoder maps $\mathbf{s}$ to subsampled speech representations $\mathbf{H} \in  \mathbb{R}^{K \times d}$, matching the codec frame rate. The causally aligned target text embeddings $\mathbf{E} \in \mathbb{R}^{K \times d}$ derived from $\mathbf{e}$ are fused with the speech embeddings $\mathbf{H}$ by element-wise addition, $\mathbf{G}=\mathbf{E}+\mathbf{H}$. The fused representation is then passed to the LLM, which produces logits $\mathbf{z}^{\text{txt}}_k$ for the next token, yielding $
P(e_k \mid e_{<k}, \mathbf{G}_{1:\kappa(k)})=\mathrm{Softmax}(\mathbf{z}^{\text{txt}}_{k}).$
The Simul-S2T objective is the autoregressive negative log-likelihood:
\begin{equation}
\mathcal{L}_{\text{s2t}}
=- \sum_{k=1}^{|\mathbf{Q}|}
\log
P(e_k \mid e_{<k}, \mathbf{G}_{1:\kappa(k)}),
\label{eq:3}
\end{equation}
where $\kappa(k)$ denotes the frame index up to which the source speech has been consumed when predicting the $k$-th target token.
For the Simul-T2S module, the translation speech waveform is first converted into discrete tokens using streaming NanoCodec. The speech decoder is a decoder-only architecture that autoregressively predicts next codec token $\mathbf{q}_k$ conditioned on previous codec tokens $\mathbf{q}_{< k}$, previous text translation $e_{\le k}$, and  fixed-dimensional speaker embedding $\mathbf{u} \in \mathbb{R}^{d}$ extracted from $s_{i}$. The Simul-T2S training objective is the autoregressive negative log-likelihood:
\begin{equation}
\mathcal{L}_{\text{t2s}}
=- \sum_{k=1}^{|\mathbf{Q}|}
\log
P(\mathbf{q}_k \mid \mathbf{q}_{<k}, e_{\leq k}, \mathbf{u}).
\label{eq:4}
\end{equation}
The overall multitask objective $\mathcal{L}_{tot}$ is the weighted sum of the two objectives:
\begin{equation}
    \mathcal{L}_{tot} = \alpha_{s2t}\mathcal{L}_{s2t} + \alpha_{t2s}\mathcal{L}_{t2s},
    \label{eq5}
\end{equation}
where $\alpha_{s2t}$ and $\alpha_{t2s}$ are hyperparameters balance the contributions of the two tasks. During training, we employ teacher forcing. At inference time, the decoder conditions on the autoregressive LLM predictions obtained via greedy decoding. Finally, the streaming codec decoder reconstructs the target waveform from the predicted discrete speech tokens.

\subsection{Causality-Aware Average Lagging (CAAL)}
\begin{figure}[tb!]
    \centering
    \vspace{-0.4cm}\includegraphics[width=1\columnwidth]{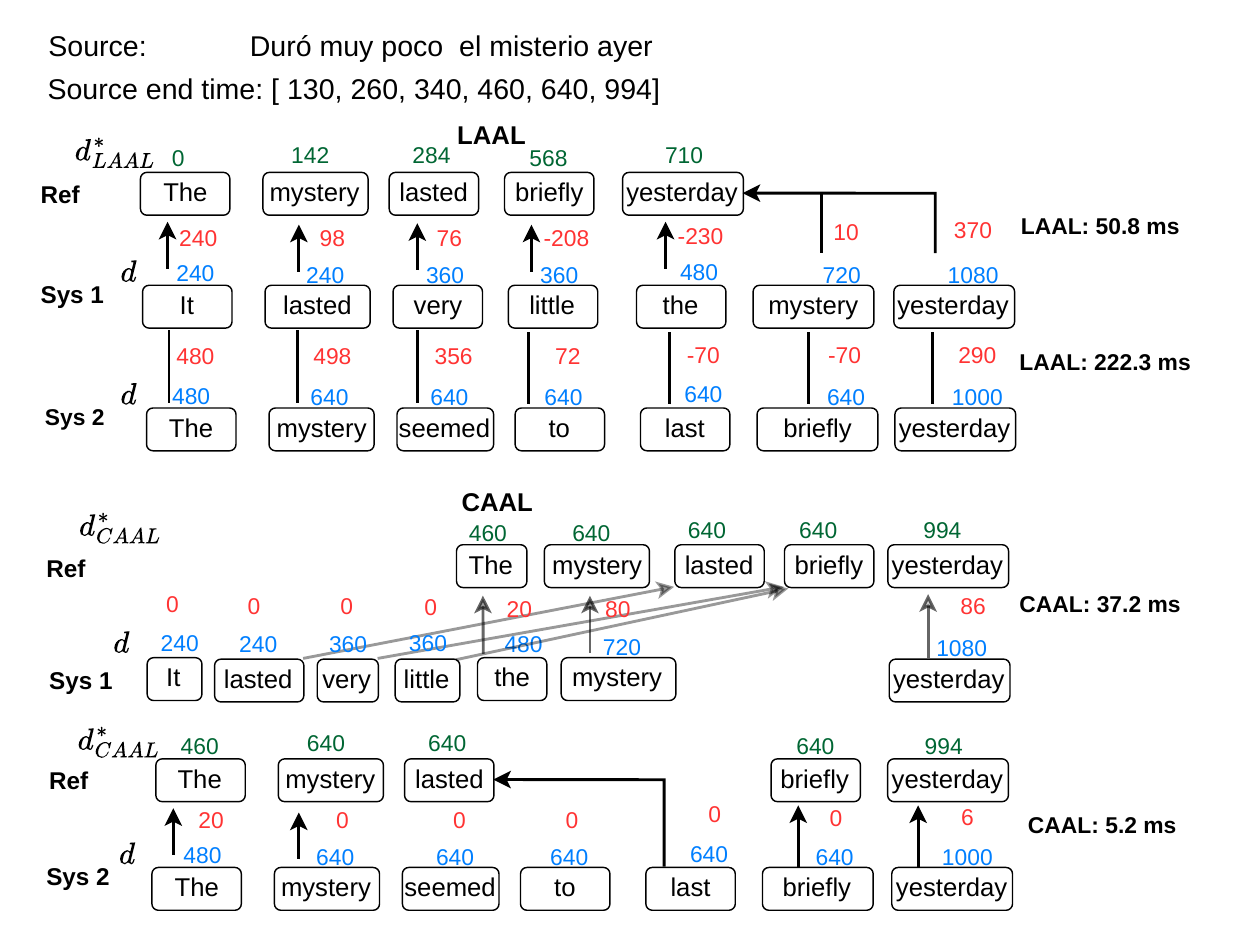}
    \caption{Illustration of LAAL and CAAL computation. Oracle delays (green) and system delays (blue) are paired according to the alignment mapping (arrows), and lagging values (red) are computed as the difference between each system delay and its aligned oracle delay.}
    \vspace{-0.4cm}
    \label{fig:maal}
\end{figure}
In this paper, we highlight key limitations of existing translation latency metrics, which do not incorporate translation alignments~\cite{cho2016can,ma-etal-2019-stacl, papi2022over, cherry2019thinking}. These metrics assume uniform word timing and penalize all delays equally, regardless of whether they are linguistically necessary. This leads to inconsistent and misleading latency estimates. We primarily compare against LAAL, a recent and widely used latency metric. Figure~\ref{fig:maal} illustrates this issue using an example from Spanish--English CVSS-T. System~1 follows a fixed wait-$k$ policy with $k=120$ ms, emitting translations word by word without modeling reordering. System~2, adopts a causality-aware dynamic chunking while consuming speech frames of 40 ms, emitting each target token only after sufficient source evidence has been observed. LAAL computes the ideal delay under a uniform timing assumption as $d_i^* = (i-1)\frac{|s^{src}_{1:T}|}{\max(|\mathbf{e}_{ref}|,|\mathbf{e}_{hyp}|)} = \frac{994}{7} \approx 142$ ms. 
Because LAAL ignores alignments, it matches oracle and system emission times incorrectly. As a result, insertions may produce negative lagging values (e.g., \textit{“little”} assigned $-208$ and \textit{“the”} assigned $-230$), which artificially lower the overall latency score. Consequently, LAAL incorrectly reports System~1 as nearly four times faster than System~2. Consequently, LAAL incorrectly reports System~1 as nearly four times faster than System~2. 

In contrast, System~2 closely follows the true source end times: it incurs only 20 ms delay for \textit{``The''}, 0 ms for \textit{``misterio''} and \textit{``Duró muy poco''}, and 6 ms for \textit{``ayer''}. Its initial 480 ms wait only slightly exceeds the 460 ms required to reorder \textit{``Duró muy poco el''} before emitting \textit{``the''}. The first 460 ms waiting is linguistically justified and should not be counted as system-induced delay. CAAL addresses this by using causal alignments to account for reordering: it neither rewards premature emissions nor penalizes necessary waiting. Instead, it isolates excess delay beyond the ideal causal policy and correctly identifies System~2 as substantially faster than System~1. 

Formally, CAAL measures the difference between the system emission time $t'_i$ for target word $e_i$ and its ideal causal delay $\tau^{en}_{src}[i]$, derived from the alignment procedure in Section~\ref{sec:data_pipeline}. It uses two alignment types:
\begin{itemize}
    \item \textbf{Causal alignment}  $\mathcal{P}$ from Section~\ref{sec:data_pipeline}
    \item \textbf{Hypothesis alignment} $\mathcal{B} \subseteq \{1,\dots,I_{\text{hyp}}\} \times \{1,\dots,I\}$, from hypothesis translation to reference translation.
\end{itemize}
Let $R$ be the set of reference word indices, and let $M\subseteq R$ denote the subset aligned to hypothesis words through $\mathcal{B}$. For aligned words, the effective delay is
\begin{equation}
d_i = \max \left(t'_i - \tau^{en}_{src}[i], 0 \right).
\end{equation}
The maximum prevents systems from artificially lowering latency by emitting reordered words before the ideal causal policy permits them. For deletions, $e_i\in R\setminus M$, we approximate the delay using the $\rho$-quantile of the sentence-level aligned empirical delay distribution:
\begin{equation}
d^{del}_i = Q_{\rho}\left(\{d_j \mid j\in M\}\right),
\label{eq:8}
\end{equation}
which penalizes omissions while remaining robust to outliers. If deletions should be ignored, one may instead set $d^{del}_i=0$.  
The CAAL metric is finally computed as
\begin{equation}
\mathrm{CAAL}
=\frac{1}{|R|}
\left(
\sum_{i\in M} d_i
+
\sum_{i\in R\setminus M} d^{del}_i
\right).
\end{equation}
Intuitively, CAAL measures the average additional waiting introduced by the system beyond an ideal causal policy that emits each target word as soon as all aligned source evidence has been observed.

\section{Experimental Setup}\label{sec:expsetup}
\textbf{Data \& pre-processing:} We evaluate Spanish, German, and French speech-to-speech translation into English. Using the pipeline in Section~\ref{sec:data_pipeline}, we construct causally aligned S2ST training data from CVSS-T and additional internal datasets, totaling approximately 2.7K hours per language pair. We extract 80-dimensional mel-spectrograms with a 25 ms window and 10 ms frame shift. During training, we apply SpecAugment with two frequency masks of width up to 27 bins and ten time masks capped at 5\% of the sequence length.\\
\begin{table}[tb!]
    \vspace{-0.4cm}
    \caption{Ablation study on the CVSS-T development set comparing alignment methods (NFA vs. MFA), multimodal representations (interleaved, IL, vs. multistream, MS), and TTS conditioning (text vs. LLM latent representations, Lat). Translation quality is evaluated with BLEU, chrF++, and COMET, with speech metrics computed on ASR transcriptions.}
    \centering

\include{tables/ablation}
    \vspace{-0.4cm}
    \label{tab:ablation}
\end{table}
\textbf{Architecture details:}
The speech encoder is initialized from a multilingual streaming FastConformer ASR model~\cite{noroozi2024stateful}, with 17 layers, 8 attention heads, 1024 attention dimension, and 2048 feed-forward dimension.  The LLM backbone is initialized from Qwen2.5-1.5B-Instruct\footnote{\url{https://huggingface.co/Qwen/Qwen2.5-1.5B-Instruct}}. The TTS component is initialized from MagpieTTS model~\cite{hussain-etal-2025-koel}, which uses streaming NanoCodec~\cite{casanova2025nanocodec} with $N_q = 13$ codebooks. The TTS decoder has 12 attention layers with dimension 768, 12 attention heads per layer, and 3072 feed-forward dimension. Both the speech encoder and the codec operate at a frame rate of 12.5 frames per second. FAST model has a total of 2B parameters. \\
\textbf{Training:}
Experiments are implemented in PyTorch with NeMo framework~\cite{kuchaiev2019nemo} and trained on 32 NVIDIA A100 (80GB) GPUs. During FAST training, all parameters are optimized except the frozen codec and speaker encoder. We use AdamW with learning rate $10^{-4}$, inverse square-root annealing with 4K warmup steps, and train for 25 epochs. The multitask weights in Eq.~(\ref{eq5}) are $\alpha_{s2t}=3$ and $\alpha_{t2s}=2$.\\
\textbf{Evaluation:} We evaluate FAST on the CVSS-T development and test sets. To ensure a comprehensive translation evaluation, we report BLEU~\cite{papineni2002bleu}, chrF++, and COMET~\cite{rei2020comet}\footnote{We use the \texttt{Unbabel/wmt22-comet-da} model.}. For speech translation evaluation, we transcribe the generated speech using a pretrained ASR\footnote{\url{https://huggingface.co/nvidia/stt_en_fastconformer_transducer_large}} and compute the same translation metrics on the resulting transcripts. All translation evaluations are case-insensitive and punctuation-free. Latency is measured with LAAL and our proposed CAAL, both implemented using the SimulEval
framework~\cite{ma2020simuleval}. For CAAL deletion penalties, we set $\rho=0.9$ in Eq.~(\ref{eq:8}), a robust high-quantile that mitigates the influence of outliers. Perceptual speech quality is evaluated with UTMOS-V2~\cite{baba2024t05}, while speaker similarity is computed as the cosine similarity between ECAPA-TDNN embeddings~\cite{desplanques2020ecapa} extracted from reference and generated speech.\\
\textbf{Inference:} During inference, the LLM decodes incrementally from speech encoder outputs arriving every 80 ms. Text predictions are generated with greedy decoding, converted to text embeddings, and passed to the TTS decoder.

\section{Results}
\subsection{Ablation analysis}\label{sec:ablation}

In this section, we assess the impact of key components of the proposed FAST approach, including alignment quality, the multimodal representation modeling, and TTS conditioning, as shown in Table~\ref{tab:ablation}. All experiments are conducted on the Spanish--English language pair using a fixed 2~s chunking policy. The baseline (Row 1) represents a common class of prior LLM-based Simul-S2T approaches that rely on interleaved speech–text multimodal representations~\cite{fu2025efficient,futami25_interspeech,ouyang2025infinisst}. In this baseline, TTS is conditioned on the text predicted by the LLM. We first assess speech-to-text alignment quality by comparing CTC-based alignments from NeMo Forced Aligner (NFA) with HMM-based  alignments from Montreal Forced Aligner (MFA). Replacing NFA with MFA improves performance by +4.4 BLEU on text and +1.0 BLEU on speech, which we attribute to MFA’s more precise word-boundary estimates. Next, we compare aligned interleaving (Row 2) and aligned multistream joint modeling (Row 4). Multistream modeling consistently outperforms interleaving, yielding gains of +3.6 BLEU, +4.9 chrF++, and +5.5 COMET on text translation, and +7.3 BLEU, +7.5 chrF++, and +5.7 COMET on speech translation. We then evaluate two TTS conditioning strategies: LLM latent representations (FAST-Lat, Row 3) and argmax-decoded text predictions (Row 4). Under our limited-data setting, latent conditioning substantially degrades speech translation quality, resulting in drops of 16.3 BLEU, 25.3 chrF++, and 12.2 COMET. Finally, to assess modularity, we replace the Qwen2-1.5B multilingual multimodal LLM with TinyLLaMA-1.2B, a text-only multilingual model. Despite the lack of multimodal pretraining, TinyLLaMA achieves comparable performance, indicating that the proposed approach is largely backbone-agnostic and does not require large-scale speech--text translation pretraining.

\begin{table}[tb!]
    
    \vspace{-0.4cm}
    \caption{Comparison of FAST on CVSS-T development set using the causality-aware adaptive policy (CAP) and a fixed policy. Translation quality is reported using BLEU, chrF++, and COMET while latency is measured using LAAL and CAAL.}
    \centering

\include{tables/translation-aware}
    \vspace{-0.4cm}
    \label{tab:trans-aware}
\end{table}

\subsection{Causality-Aware Adaptive Policy}\label{sec:adaptive-policy}
\begin{table*}[tb!]
    \vspace{-0.4cm}
    \caption{Comparison of translation performance on the CVSS-T test set. Translation quality is evaluated with BLEU, chrF++, and COMET, with speech metrics computed on ASR transcriptions. Audio quality is measured using UTMOS-V2, and speaker similarity is measured using cosine similarity.}
    \centering
\include{tables/multilingual}
    \vspace{-0.4cm}
    \label{tab:multilingual}
\end{table*}
In this section, we evaluate whether the proposed causality-aware adaptive policy (CAP) improves the quality–latency trade-off over the commonly used fixed chunking policy adopted by several prior LLM-based Simul-S2T approaches~\cite{agostinelli2024simul,koshkin2024llms,futami25_interspeech,ouyang2025infinisst}. In all experiments, we use the best-performing FAST configuration from Table~\ref{tab:ablation}. Without chunk merging (0.4~s average input), performance degrades substantially despite sufficient source information being theoretically available to the LLM. Increasing the average input chunk duration to 0.9~s through chunk merging improves text translation quality to near-baseline levels, suggesting that the LLM benefits from additional contextual grounding. At an average input duration of 1.5~s, FAST-CAP surpasses the fixed 2~s policy, improving text translation by +1.1 BLEU, +3.0 chrF++, and +2.5 COMET. 

Next, we analyze latency using both LAAL and the proposed CAAL metric. As expected, latency increases with larger input chunks; however, LAAL becomes less discriminative at longer chunk durations. For example, on Spanish--English, FAST-CAP (1.5~s) and FAST-Fixed (2~s) differ by only 1.7\% in LAAL, whereas CAAL captures a more substantial 8\% relative reduction. This suggests that CAAL provides a more discriminative measure of latency. To evaluate the generalization of FAST-CAP beyond Spanish, we extend the comparison to German and French. On German, FAST-CAP improves text translation by +1.2 BLEU, +2.8 chrF++, and +0.6 COMET, while reducing CAAL by 16\% relative, compared to only 4.7\% under LAAL. On French, translation performance remains comparable to the fixed baseline, while FAST-CAP achieves a 26\% relative reduction in CAAL, compared to 16.6\% under LAAL. These results demonstrate that FAST-CAP consistently improves the quality--latency trade-off across language pairs. Moreover, CAAL provides a more discriminative assessment of latency than LAAL.


\subsection{Comparison with state-of-the-art models}\label{sec:multiling}
In this section, we compare FAST-CAP with Hibiki-Zero~\cite{labiausse2026simultaneous} and streaming SeamlessM4T~\cite{barrault2023seamless}, as shown in Table~\ref{tab:multilingual}.
 To better match their scale, we replace the 0.1B multilingual ASR encoder with a 0.6B encoder\footnote{\url{https://huggingface.co/nvidia/nemotron-3.5-asr-streaming-0.6b}}, yielding a 2.5B-parameter model, and increase the average input chunk size from 1.5~s to 2~s. FAST-CAP is trained on  total of 8K hours of multilingual data across French, Spanish, and German, compared with 160K hours for Hibiki-Zero and 145K/245K hours of speech-to-speech/speech-to-text data for SeamlessM4T. Despite this data gap, FAST-CAP outperforms Hibiki-Zero in text translation quality, improving by +2.3 BLEU, +1.3 chrF++, and +3.2 COMET, while achieving comparable speaker similarity and reducing latency by 38.8\% in CAAL and 40.8\% in LAAL. Compared with SeamlessM4T, FAST-CAP achieves comparable text translation and audio quality, higher speaker similarity (0.44 vs. 0.31), and relative latency reductions of 30.9\% in CAAL and 33.9\% in LAAL. In both comparisons, the main remaining gap for FAST-CAP lies in generated speech translation quality. To improve generated speech quality, FAST-CAP-L adopts the TTS model from Audio Flamingo 3-Chat~\cite{ghosh2026audio}, which clones the speaker’s voice using an audio prompt. FAST-CAP-L also uses a larger translation-focused LLM\footnote{\url{https://huggingface.co/nvidia/Riva-Translate-4B-Instruct-v1.1}} to broaden language coverage, bringing the total model size to 5B parameters. FAST-CAP-L retains FAST-CAP’s latency while achieving the best speech translation scores across BLEU, chrF++, and COMET, with the largest gain in COMET: +2.1 points above Hibiki-Zero. It also achieves the highest speaker similarity (0.53 versus 0.47 for Hibiki-Zero) while maintaining competitive audio quality.

\section{Conclusion}
In this work, we introduced a causality-aware framework for LLM-based simultaneous speech-to-speech translation that integrates causal constraints across data generation, source--target alignment, translation policy, and latency evaluation. Our novel data pipeline derives a target-driven adaptive policy that generates each target segment once sufficient source context is available. 
The factorized S2ST architecture (FAST) decouples perception and generation representations, improving translation accuracy without compromising synthesis fidelity. We also proposed an alignment-aware latency metric (CAAL) that separates linguistically necessary reordering from avoidable system-induced delay. Experiments on multilingual benchmarks show that FAST combined with the causality-aware adaptive policy (CAP) provides better quality--latency trade-offs than fixed-policy baselines. Despite using limited training data, FAST-CAP achieves state-of-the-art speech translation quality while providing lower latency and higher speaker fidelity. CAAL also offers a more accurate and discriminative latency assessment than existing latency metrics. Future work will focus on further improving cross-lingual voice transfer, extending the framework to longer conversational contexts, and supporting multi-party and multimodal interactions.

\newpage
\section*{References}





\printbibliography
\end{document}

%% file: sourcemap_definitions.tex
\DeclareSourcemap{
  \maps[datatype=bibtex, overwrite=true]{
    \map{
      \step[fieldsource=booktitle, match=\regexp{.*Interspeech.*}, replace={Proc. Interspeech}]
      \step[fieldsource=journal, match=\regexp{.*INTERSPEECH.*}, replace={Proc. Interspeech}]
      \step[fieldsource=booktitle, match=\regexp{.*ICASSP.*}, replace={Proc. ICASSP}]
      \step[fieldsource=booktitle, match=\regexp{.*icassp_inpress.*}, replace={Proc. ICASSP (in press)}]
      \step[fieldsource=booktitle, match=\regexp{.*Acoustics,.*Speech.*and.*Signal.*Processing.*}, replace={Proc. ICASSP}]
      \step[fieldsource=booktitle, match=\regexp{.*International.*Conference.*on.*Learning.*Representations.*}, replace={Proc. ICLR}]
      \step[fieldsource=booktitle, match=\regexp{.*International.*Conference.*on.*Computational.*Linguistics.*}, replace={Proc. COLING}]
      \step[fieldsource=booktitle, match=\regexp{.*SIGdial.*Meeting.*on.*Discourse.*and.*Dialogue.*}, replace={Proc. SIGDIAL}]
      \step[fieldsource=booktitle, match=\regexp{.*International.*Conference.*on.*Machine.*Learning.*}, replace={Proc. ICML}]
      \step[fieldsource=booktitle, match=\regexp{.*North.*American.*Chapter.*of.*the.*Association.*for.*Computational.*Linguistics:.*Human.*Language.*Technologies.*}, replace={Proc. NAACL}]
      \step[fieldsource=booktitle, match=\regexp{.*Empirical.*Methods.*in.*Natural.*Language.*Processing.*}, replace={Proc. EMNLP}]
      \step[fieldsource=booktitle, match=\regexp{.*Association.*for.*Computational.*Linguistics.*}, replace={Proc. ACL}]
      \step[fieldsource=booktitle, match=\regexp{.*Automatic.*Speech.*Recognition.*and.*Understanding.*}, replace={Proc. ASRU}]
      \step[fieldsource=booktitle, match=\regexp{.*Spoken.*Language.*Technology.*}, replace={Proc. SLT}]
      \step[fieldsource=booktitle, match=\regexp{.*Speech.*Synthesis.*Workshop.*}, replace={Proc. SSW}]
      \step[fieldsource=booktitle, match=\regexp{.*workshop.*on.*speech.*synthesis.*}, replace={Proc. SSW}]
      \step[fieldsource=booktitle, match=\regexp{.*Advances.*in.*neural.*information.*processing.*}, replace={Proc. NeurIPS}]
      \step[fieldsource=booktitle, match=\regexp{.*Advances.*in.*Neural.*Information.*Processing.*}, replace={Proc. NeurIPS}]
      \step[fieldsource=booktitle, match=\regexp{.*Workshop.*on.*Applications.*of.*Signal.*Processing.*to.*Audio.*and.*Acoustics.*}, replace={Proc. WASPAA}]
      \step[fieldsource=publisher, match=\regexp{.+}, replace={{}}]
      \step[fieldsource=month, match=\regexp{.+}, replace={{}}]
      \step[fieldsource=location, match=\regexp{.+}, replace={{}}]
      \step[fieldsource=address, match=\regexp{.+}, replace={{}}]
      \step[fieldsource=organization, match=\regexp{.+}, replace={{}}]
    }
  }
}

%% file: tables/data.tex
\setlength{\tabcolsep}{2pt}
\resizebox{0.48\textwidth}{!}{
\begin{tabular}{l c c c c c c}
\toprule
     & \multicolumn{2}{c}{\textbf{Es-En}} 
& \multicolumn{2}{c}{\textbf{De-En}} & \multicolumn{2}{c}{\textbf{Fr-En}} \\
\cmidrule(lr){2-3} \cmidrule(lr){4-5} \cmidrule(lr){6-7}
& \textbf{UTMOS} & \textbf{Spk-Sim} & \textbf{UTMOS} & \textbf{Spk-Sim} & \textbf{UTMOS} &\textbf{Spk-Sim} \\
\midrule
CVSS-T (public)  & 2.4 & 0.22 & 2.5 & 0.26 & 2.5 & 0.25 \\
CVSS-T (ours)  & 2.4 & \textbf{0.64} & 2.5 & \textbf{0.61} & 2.5 & \textbf{0.57}  \\
\bottomrule
\end{tabular}
}

%% file: tables/ablation.tex
\setlength{\tabcolsep}{3pt}
\resizebox{0.49\textwidth}{!}{
\begin{tabular}{l l c c c c c c c c}
\toprule
& & & 
& \multicolumn{3}{c}{\textbf{Text Translation}} 
& \multicolumn{3}{c}{\textbf{ASR Transcript}} \\
\cmidrule(lr){5-7} \cmidrule(lr){8-10}
\textbf{\#} & \textbf{System} & \textbf{LLM} & \textbf{Align.}
& \textbf{BLEU} & \textbf{chrF} & \textbf{COMET}
& \textbf{BLEU} & \textbf{chrF} & \textbf{COMET} \\
\midrule

1 & FAST (IL) & Qwen & NFA & 18.7 & 49.0 & 68.6 & 12.5 & 39.5 & 61.1 \\
2 & FAST (IL) & Qwen & MFA & \textbf{23.1} & \textbf{49.7} & \textbf{70.0} & \textbf{13.5} & \textbf{40.2} & \textbf{62.5} \\

\midrule
3 & FAST-Lat (MS) & Qwen & MFA & 26.6 & 54.8 & 75.4 & 4.5 & 23.3 & 56.0 \\
4 & FAST (MS) & Qwen & MFA & \textbf{26.7} & \textbf{54.6} & \textbf{75.5} & \textbf{20.8} & \textbf{48.6} & \textbf{68.2} \\

\midrule
5 & FAST (MS) & Llama & MFA & 25.1 & 54.3 & 75.4 & 19.9 & 48.0 & 67.5 \\

\bottomrule
\end{tabular}
}

%% file: tables/translation-aware.tex
\setlength{\tabcolsep}{3pt}
\resizebox{0.49\textwidth}{!}{
\begin{tabular}{l c c c c c c}
\toprule
& \textbf{Avg.} 
& \multicolumn{2}{c}{\textbf{Latency (s)}} 
& \multicolumn{3}{c}{\textbf{Translation Quality}} \\
\cmidrule(lr){3-4} \cmidrule(lr){5-7}
\textbf{System} 
& \textbf{input (s)}
& \textbf{CAAL $\downarrow$} 
& \textbf{LAAL $\downarrow$}
& \textbf{BLEU $\uparrow$} 
& \textbf{chrF $\uparrow$} 
& \textbf{COMET $\uparrow$} \\
\midrule

FAST-CAP (Sp-En) & 0.4 & 0.33 & 0.57 & 12.1 & 47.0 & 67.1 \\
FAST-CAP (Sp-En) & 0.9 & 0.52 & 0.88 & 24.0 & 55.3 & 75.0 \\

FAST-CAP (Sp-En) & \textbf{1.5} & \textbf{0.80} & \textbf{1.15} & \textbf{27.8} & \textbf{57.6} & \textbf{78.0} \\
FAST-Fixed (Sp-En) & 2.0 & 0.87 & 1.17 & 26.7 & 54.6 & 75.5 \\

\midrule
FAST-CAP (De-En) & \textbf{1.5} & \textbf{0.68} & \textbf{1.20} & \textbf{34.6} & \textbf{62.0} & \textbf{78.1} \\
FAST-Fixed (De-En) & 2.0 & 0.81 & 1.26 & 33.4 & 59.2 & 77.5 \\

\midrule
FAST-CAP (Fr-En) & \textbf{1.5} & \textbf{0.67} & \textbf{1.00} & \textbf{32.8} & \textbf{60.8} & \textbf{78.3} \\
FAST-Fixed (Fr-En) & 2.0 & 0.91 & 1.20 & 32.7 & 60.7 & 78.0 \\

\bottomrule
\end{tabular}
}

%% file: tables/multilingual.tex
\setlength{\tabcolsep}{3pt}
\resizebox{0.8\textwidth}{!}{
\begin{tabular}{l c c c c c c c c c c}
\toprule
& \multicolumn{3}{c}{\textbf{Text Translation}} 
& \multicolumn{3}{c}{\textbf{ASR Transcript}} 
& \multicolumn{2}{c}{\textbf{Latency}} & \multicolumn{2}{c}{\textbf{Speech Quality}} \\
\cmidrule(lr){2-4} \cmidrule(lr){5-7} \cmidrule(lr){8-9} \cmidrule(lr){10-11}
\textbf{System} 
& \textbf{BLEU} $\uparrow$& \textbf{chrF} $\uparrow$& \textbf{COMET}$\uparrow$
& \textbf{BLEU} $\uparrow$& \textbf{chrF} $\uparrow$& \textbf{COMET}$\uparrow$
& \textbf{CAAL} $\downarrow$& \textbf{LAAL} $\downarrow$& \textbf{UTMOS-V2}$\uparrow$ & \textbf{SpkSim} $\uparrow$\\
\midrule
Hibiki-Zero
& 32.9 & 59.9 & 75.8
& 31.4 & 58.2 &74.9
& 1.39 & 2.11 & \textbf{2.1} & 0.47 \\

SeamlessM4T  &
\textbf{36.0} & \textbf{63.1} & 79.6
& 32.2 & 58.2 & 74.7
& 1.23 & 1.89 & 1.9 & 0.31 \\
FAST-CAP 
& 35.2 & 61.2 & 79.0
& 27.6 & 54.3 & 70.5
& \textbf{0.85} & \textbf{1.25} & 1.8 & 0.44 \\
FAST-CAP-L 
& 35.8 & 62.1 & \textbf{80.6}
& \textbf{32.8} & \textbf{58.6} & \textbf{77.0}
& \textbf{0.85} & \textbf{1.25} & 2.0 & \textbf{0.53} \\
\bottomrule
\end{tabular}}